\documentclass{article} 

     \PassOptionsToPackage{square, numbers}{natbib}

     \usepackage[preprint]{neurips_2018}
     
\usepackage[utf8]{inputenc} 
\usepackage[T1]{fontenc}    
\usepackage{hyperref}       
\usepackage{url}            
\usepackage{booktabs}       
\usepackage{amsfonts}       
\usepackage{nicefrac}       
\usepackage{microtype}      

\usepackage[]{algorithm}
\usepackage{algpseudocode}
\usepackage{amsmath}
\usepackage{graphicx}
\usepackage{epstopdf}
\usepackage{subcaption}
\usepackage{enumitem} 

\title{Layer rotation: a surprisingly powerful indicator of generalization in deep networks?}

\author{Simon Carbonnelle, Christophe De Vleeschouwer \\
FNRS research fellows\\
ICTEAM, Université catholique de Louvain\\
Louvain-La-Neuve, Belgium \\
\texttt{\{simon.carbonnelle, christophe.devleeschouwer\}@uclouvain.be} 
}

\begin{document}

\maketitle

\begin{abstract}
Our work presents extensive empirical evidence that layer rotation, i.e. the evolution across training of the cosine distance between each layer's weight vector and its initialization, constitutes an impressively consistent indicator of generalization performance. In particular, larger cosine distances between final and initial weights of each layer consistently translate into better generalization performance of the final model. Interestingly, this relation admits a network independent optimum: training procedures during which all layers' weights reach a cosine distance of 1 from their initialization consistently outperform other configurations -by up to $30\%$ test accuracy. Moreover, we show that layer rotations are easily monitored and controlled (helpful for hyperparameter tuning) and potentially provide a unified framework to explain the impact of learning rate tuning, weight decay, learning rate warmups and adaptive gradient methods on generalization and training speed. In an attempt to explain the surprising properties of layer rotation, we show on a 1-layer MLP trained on MNIST that layer rotation correlates with the degree to which features of intermediate layers have been trained. 
\end{abstract}

\section{Introduction}
%
%
In order to understand the intriguing generalization properties of deep neural networks highlighted by \cite{Neyshabur2015,Zhang2017,Keskar2017}, the identification of numerical indicators of generalization performance that remain applicable across a diverse set of training settings is critical. A well-known and extensively studied example of such indicator is the width of the minima the network has converged to \citep{Hochreiter1997a,Keskar2017}. These indicators provide important insights for theoretical works around generalization in deep learning, and help explaining why commonly used tricks and techniques influence generalization performance.

In this paper, we present empirical evidence supporting the discovery of a novel indicator of generalization: \textbf{the evolution across training of the cosine distance between each layer's weight vector and its initialization} (denoted by \textit{layer rotation}). Indeed, we show across a diverse set of experiments (with varying datasets, networks and training procedures), that larger layer rotations (i.e. larger cosine distance between final and initial weights of each layer) consistently translate into better generalization performance. In addition to providing an original perspective on generalization, our experiments suggest that layer rotation also benefits from the following properties compared to alternative indicators of generalization:
\begin{itemize}
\item It has a network-independent optimum (all layers reaching a cosine distance of 1)
\item It is easily monitored and, since it only depends on the evolution of the network's weights, can be controlled along the optimization through appropriate weight update adjustments
\item It provides a unified framework to explain the impact of learning rate tuning, weight decay, learning rate warmups and adaptive gradient methods on generalization and training speed.
\end{itemize}
In comparison, other indicators usually provide a metric to optimize (\textit{e.g.} the wider the minimum, the better) but no clear optimum to be reached (what is the optimal width?), nor a precise methodology to tune it (how to converge to a minimum with a specific width?). By disclosing simple guidelines to tune layer rotations and an easy-to-use controlling tool, our work can also help practitioners get the best out of their network with minimal hyper-parameter tuning.

After a discussion of related work (Section \ref{sec:related}), the presentation of our experimental study is structured according to three successive steps: 
\begin{enumerate}
\item Development of tools to monitor and control layer rotation (Section \ref{sec:tools});
\item Systematic study of layer rotation configurations in a controlled setting (Section \ref{sec:Exploration});\footnotemark
\item Study of layer rotation configurations in standard training settings, with a special focus on SGD, weight decay and adaptive gradient methods (Section \ref{sec:CommonPracticesAnalysis}).\footnotemark[1]
\end{enumerate}
\footnotetext{Our study focuses on convolutional neural networks used for image classification.}
Finally, Section \ref{sec:future} provides preliminary results and discussion for explaining the surprising properties of layer rotation.

To encourage further validation of our claims, the tools and source code used to create all the figures of this paper are provided at \url{https://github.com/ispgroupucl/layer-rotation-paper-experiments} (code uses the Keras \citep{chollet2015keras} and TensorFlow \citep{Agarwal2016} libraries). In order to facilitate the usage of our controlling and monitoring tool by practitioners, implementations in different deep learning libraries are available at \url{https://github.com/ispgroupucl/layer-rotation-tools}.

\section{Related work} \label{sec:related}
After the intriguing generalization properties of deep neural networks were highlighted by \cite{Neyshabur2015,Zhang2017,Keskar2017}, several works have tried to identify aspects of training which could predict generalization performance in a consistent and general way. 

A first line of work tries to identify the characteristics of \textit{trained models} that correlate with good generalization properties. Different complexity metrics have been proposed: norm-based metrics are studied in \citep{Bartlett1998,Neyshabur2015a,Liang2019}, sensitivity-based metrics in \citep{Xu2012,Novak2018} and sharpness-based metrics in \citep{Hochreiter1997a,Keskar2017}. These three approaches are compared and further studied in \citep{Neyshabur2017}. Other works proposed complexity metrics that explicitly use the decomposition of deep neural networks in layers. Such decomposition is also key to our work and has already been very successful when analysing the training difficulties of deep neural networks \citep{Bengio1994,Hochreiter1998,Glorot2010}. In \citep{Morcos2018}, the similarity of hidden representations resulting from training with different initializations is used as an indicator of generalization. In \citep{Morcos2018a}, sensitivity to perturbation of hidden representations is studied. 

While the works described above help us understand the characteristics of models that generalize well, they don't explicitly disclose how \textit{the training procedure} itself leads to such characteristics. A second line of work, to which this paper belongs, studies indicators of generalization that characterize the whole training trajectory instead of solely focusing on its endpoint. The sharpness metric is revisited in \citep{Jastrzebski2019} and \citep{Fort} analyses stiffness. While the presence of noise is believed to affect generalization, the mechanisms at play are poorly understood \cite{Goodfellow2015,Xing2018} and a clear metric to quantify its influence on generalization is still lacking. \cite{Hoffer2017} studies the evolution of the euclidean distance between the model's weight vector and its initialization in the context of large batch training. This metric is the most similar to layer rotation, and also has the particularity of being both easy to monitor and to control. Our work differs by the used distance metric (layer-level cosine distance instead of model-level euclidean distance) and by performing a larger study that extends the context of large-batch training.

\section{Tools for monitoring and controlling layer rotation} \label{sec:tools}
This section describes the tools for monitoring and controlling layer rotation during training, such as its relation with generalization can be studied in Sections \ref{sec:Exploration} and \ref{sec:CommonPracticesAnalysis}.

\subsection{Monitoring layer rotation with layer rotation curves} \label{sec:monitoring}
Layer rotation is defined as the evolution of the cosine distance between each layer's weight vector and its initialization during training. More precisely, let $w_l^t$ be the flattened weight tensor of the $l^{th}$ layer at optimization step $t$ ($t_0$ corresponding to initialization), then the rotation of layer $l$ at training step $t$ is defined as the cosine distance between $w_l^{t_0}$ and $w_l^{t}$. \footnote{It is worth noting that our study focuses on weights that multiply the inputs of a layer (\textit{e.g.} kernels of fully connected and convolutional layers). Studying the training of additive weights (biases) is left as future work.} In order to visualize the evolution of layer rotation during training, we record how the cosine distance between each layer's current weight vector and its initialization evolves across training steps. We denote this visualization tool by \textit{layer rotation curves} hereafter.
 
\subsection{Controlling layer rotation with Layca} \label{sec:layca}
The ability to control layer rotations during training would enable a systematic study of its relation with generalization. Therefore, we present Layca (LAYer-level Controlled Amount of weight rotation), an algorithm where the layer-wise learning rates directly determine the amount of rotation performed by each layer's weight vector during each optimization step (the \textit{layer rotation rates}), in a direction specified by an optimizer (SGD being the default choice). Inspired by techniques for optimization on manifolds \citep{absil2010}, and on spheres in particular, Layca applies layer-wise orthogonal projection and normalization operations on SGD's updates, as detailed in Algorithm \ref{alg:layca} in Supplementary Material. These operations induce the following simple relation between the learning rate $\rho_l (t)$ of layer $l$ at training step $t$ and the angle $\theta_l (t)$ between $w_l^t$ and $w_l^{t-1}$: $\rho_l (t) = tan(\theta_l (t))$. 

Our controlling tool is based on a strong assumption: that controlling the amount of rotation performed during each individual training step (i.e. the layer rotation rate) enables control of the cumulative amount of rotation performed since the start of training (i.e. layer rotation). This assumption is not trivial since the aggregated rotation is a priori very dependent on the structure of the loss landscape. For example, for an identical layer rotation rate, the layer rotation will be much smaller if iterates oscillate around a minimum instead of following a stable downward slope. As will be attested by the inspection of the layer rotation curves, our assumption however appeared to be sufficiently valid, and the control of layer rotation was effective in our experiments.

\begin{algorithm*}
   \caption{Layca, an algorithm that enables control over the amount of weight rotation per step for each layer through its learning rate parameter (cfr. Section \ref{sec:layca}).}
   \label{alg:layca}
\begin{algorithmic}
   \State {\bfseries Require:} $o$, an optimizer (SGD is the default choice)
   \State {\bfseries Require:} $T$, the number of training steps
   \State L is the number of layers in the network
   \For{l=0 {\bfseries to} L-1}
   \State {\bfseries Require:} $\rho _l (t)$, a layer's learning rate schedule
   \State {\bfseries Require:} $w^l_0$, the initial multiplicative weights of layer $l$
   \EndFor
   \State $t \leftarrow 0$
   \While{$t<T$}
   \State $s^0_t,..., s^{L-1}_t  = \text{getStep}(o,w^0_t,..., w^{L-1}_t)$ \noindent\hspace{6pt} (get the updates of the selected optimizer)
   \For{l=0 {\bfseries to} L-1}
   \State $s^l_t \leftarrow s^l_t - \frac{(s^l_t \cdot w^l_t) w^l_t}{w^l_t \cdot w^l_t}$ \noindent\hspace{24pt} (project step on space orthogonal to $w^l_t$)
   \State $s^l_t \leftarrow \frac{s^l_t \parallel w^l_t\parallel _2}{\parallel s^l_t\parallel _2}$                 \noindent\hspace{50pt} (rotation-based normalization)
   \State $w^l_{t+1} \leftarrow w^l_t + \rho_l(t) s^l_t$ \noindent\hspace{18pt} (perform update)
   \State $w^l_{t+1} \leftarrow w^l_{t+1} \frac{\parallel w^l_{0}\parallel _2}{\parallel w^l_{t+1}\parallel _2}$ \noindent\hspace{16pt} (project weights back on sphere)
   \EndFor
   \State $t \leftarrow t+1$
   \EndWhile
\end{algorithmic}
\end{algorithm*}

\section{A systematic study of layer rotation configurations with Layca} \label{sec:Exploration}
Section \ref{sec:tools} provides tools to monitor and control layer rotation. The purpose of this section is to use these tools to conduct a systematic experimental study of layer rotation configurations. We adopt SGD as default optimizer, but use Layca (cfr. Algorithm \ref{alg:layca}) to vary the relative rotation rates (faster rotation for first layers, last layers, or no prioritization) and the global rotation rate value (high or low rate, for all layers). The experiments are conducted on five different tasks which vary in network architecture and dataset complexity, and are further described in Table \ref{tab:experiments}.

\begin{table*}[!h]
  \caption{Summary of the tasks used for our experiments\protect\footnotemark}
  \label{tab:experiments}
  \centering
  \begin{tabular}{lll}
    \toprule
    Name     & Architecture     & Dataset \\
    \midrule
    C10-CNN1     & VGG-style 25 layers deep CNN  & CIFAR-10    \\
    C100-resnet  & ResNet-32  & CIFAR-100      \\
    tiny-CNN     & VGG-style 11 layers deep CNN       & Tiny ImageNet  \\
    C10-CNN2     & deep CNN from torch blog         & CIFAR-10 + data augm.  \\
    C100-WRN     & Wide ResNet 28-10 with 0.3 dropout        & CIFAR-100 + data augm.  \\
    \bottomrule
  \end{tabular}
\end{table*}
\footnotetext{References: VGG \citep{Simonyan2014}, ResNet \citep{He2016}, torch blog \citep{torchCNN}, Wide ResNet \citep{Zagoruyko2016}, CIFAR-10 \citep{Krizhevsky2009}, Tiny ImageNet \citep{Deng2009,CS231N}. Dropout layers were removed from the torch blog CNN to enable perfect classification on the training set ($100\%$ accuracy).}

\subsection{Layer rotation rate configurations}
Layca enables us to specify layer rotation rate configurations, i.e. the amount of rotation performed by each layer's weight vector during one optimization step, by setting the layer-wise learning rates. To explore the large space of possible layer rotation rate configurations, our study restricts itself to two directions of variation. First, we vary the initial global learning rate $\rho (0)$, which affects the layer rotation rate of all the layers. During training, the global learning rate $\rho (t)$ drops following a fixed decay scheme (hence the dependence on $t$), as is common in the literature (cfr. Supplementary Material \ref{sec:lrDecay}). The second direction of variation tunes the relative rotation rates between different layers. More precisely, we apply static, layer-wise learning rate multipliers that exponentially increase/decrease with layer depth (which is typically encountered with exploding/vanishing gradients). The multipliers are parametrized by the layer index $l$ (in forward pass ordering) and a parameter $\alpha \in [-1,1]$ such that the learning rate of layer $l$ becomes:
\begin{equation} \label{eq:alpha}
\rho_l(t) =
\left\lbrace
\begin{array}{lll}
(1-\alpha)^{5\frac{(L-1-l)}{L-1}} \rho (t)  & \mbox{if} & \alpha>0\\
(1+\alpha)^{5\frac{l}{L-1}} \rho (t) & \mbox{if} & \alpha \leq 0
\end{array}\right.
\end{equation}
Values of $\alpha$ close to $-1$ correspond to faster rotation of first layers, $0$ corresponds to uniform rotation rates, and values close to $1$ to faster rotation of last layers. Visualization of the layer-wise multipliers for different $\alpha$ values is provided in Supplementary Material (\ref{sec:alphaviz}). 

We explore 10 logarithmically spaced values of $\rho (0)$ ($3^{-7},3^{-6},...,3^2$) in the $\alpha = 0$ setting, and 13 different values of $\alpha$ in the $\rho (0) = 3^{-3}$ setting. A lower amount of configurations ($\alpha \in \{-0.6, 0., 0.6\}$ and $\rho (0)\in \{3^{-5},3^{-4},3^{-3}\}$) are investigated for the C10-CNN2 and C100-WRN tasks given the increased computational cost required for their training.

\subsection{Study of the relation between layer rotation and generalization} \label{subsec:generalization}
\begin{figure*}[!h]
\begin{center}
\includegraphics[width=0.92\linewidth]{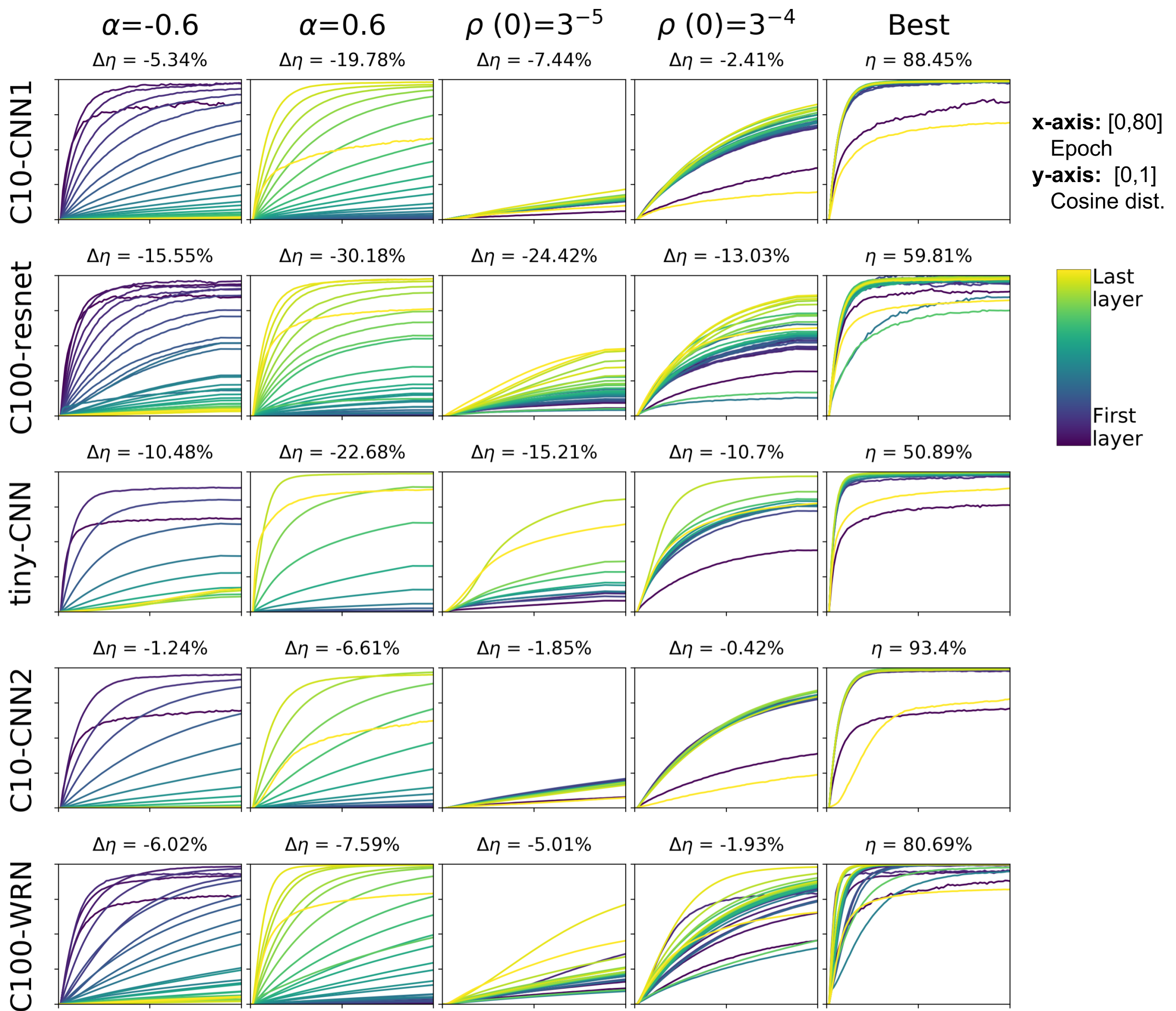}
\end{center}
\caption{Analysis of the layer rotation curves (cfr. Section \ref{sec:monitoring}) and test accuracies ($\eta$) induced by different layer rotation rate configurations (using Layca for training) on the five tasks of Table \ref{tab:experiments}. The configurations are parametrized by $\alpha$, that controls which layers have the highest rotation rates (first layers for $\alpha <0$, last layers for $\alpha >0$, or no prioritization for $\alpha=0$), and $\rho (0)$, the initial global rotation rate value shared by all layers. $\Delta\eta$ is computed with respect to the best configuration (last column), which corresponds to $\alpha = 0$ and $\rho (0) = 3^{-3}$ for the five tasks. This visualization unveils large differences in generalization ability across configurations which seem to follow a simple yet consistent rule of thumb: the larger the layer rotation for each layer, the better the generalization performance. Training accuracies are provided in Supplementary Material ($\approx 100\% $ in all configurations).}
\label{fig:exploration_curves}
\end{figure*}

Figure \ref{fig:exploration_curves} depicts the layer rotation curves (cfr. Section \ref{sec:monitoring}) and the corresponding test accuracies obtained with different layer rotation rate configurations (results for a larger set of $\alpha$ and $\rho (0)$ configurations are provided in Supplementary Material (Figure \ref{fig:exploration_accs})). While each configuration solves the classification task on the training data ($\approx 100\%$ training accuracy in all configurations, cfr. Supplementary Material \ref{sec:trainingerrors}), we observe huge differences in generalization ability (differences of up to $30\%$ test accuracy). More importantly, these differences in generalization ability seem to be tightly connected to differences in layer rotations. In particular, we extract the following rule of thumb that is applicable across the five considered tasks: the larger the layer rotations, the better the generalization performance. The best performance is consistently obtained when nearly all layers reach the largest possible distance from their initialization: a cosine distance of $1$ (cfr. fifth column of Figure \ref{fig:exploration_curves}). This observation would have limited value if many configurations (amongst which the best one) lead to cosine distances of $1$. However, we notice that most configurations do not. In particular, rotating the layers weights very slightly is sufficient for the network to achieve $100\%$ training accuracy (cfr. third column of Figure \ref{fig:exploration_curves}).

We also observe that layer rotation rates (rotation with respect to the previous iterate) translate remarkably well in layer rotations (rotation with respect to the initialization). For example, the $\alpha = 0$ setting used in the fifth column indeed leads all layers to rotate quasi synchronously. As discussed in Section \ref{sec:layca}, this is not self-evident. Understanding why this happens (and why the first and last layers seem to be less tameable) is an interesting direction of research resulting from our work.

\newpage
\section{A study of layer rotation in standard training settings} \label{sec:CommonPracticesAnalysis}
Section \ref{sec:Exploration} uses Layca to study the relation between layer rotations and generalization in a controlled setting. This section investigates the layer rotation configurations that naturally emerge when using SGD, weight decay or adaptive gradient methods for training. First of all, these experiments will provide supplementary evidence for the rule of thumb proposed in Section \ref{sec:Exploration}. Second, we'll see that studying training methods from the perspective of layer rotation can provide useful insights to explain their behaviour.

The experiments are performed on the five tasks of Table \ref{tab:experiments}. The learning rate parameter is tuned independently for each training setting through grid search over 10 logarithmically spaced values ($3^{-7},3^{-6},...,3^2$), except for C10-CNN2 and C100-WRN where learning rates are taken from their original implementations when using SGD + weight decay, and from \citep{Wilson2017} when using adaptive gradient methods for training. The test accuracies obtained in standard settings will often be compared to the best results obtained with Layca, which are provided in the 5th column of Figure \ref{fig:exploration_curves}.

\subsection{Analysis of SGD's learning rate} \label{sec:SGDanalysis}
The influence of SGD's learning rate on generalization has been highlighted by several works \citep{Jastrz2017,SmithSam2017,Smith2017,Hoffer2017,masters2018revisiting}. The learning rate parameter directly affects layer rotation rates, since it changes the size of the updates. In this section, we verify if the learning rate's impact on generalization is coherent with our rules of thumb around layer rotation.

\begin{figure}[!h]
\begin{center}
\includegraphics[width=.92\linewidth]{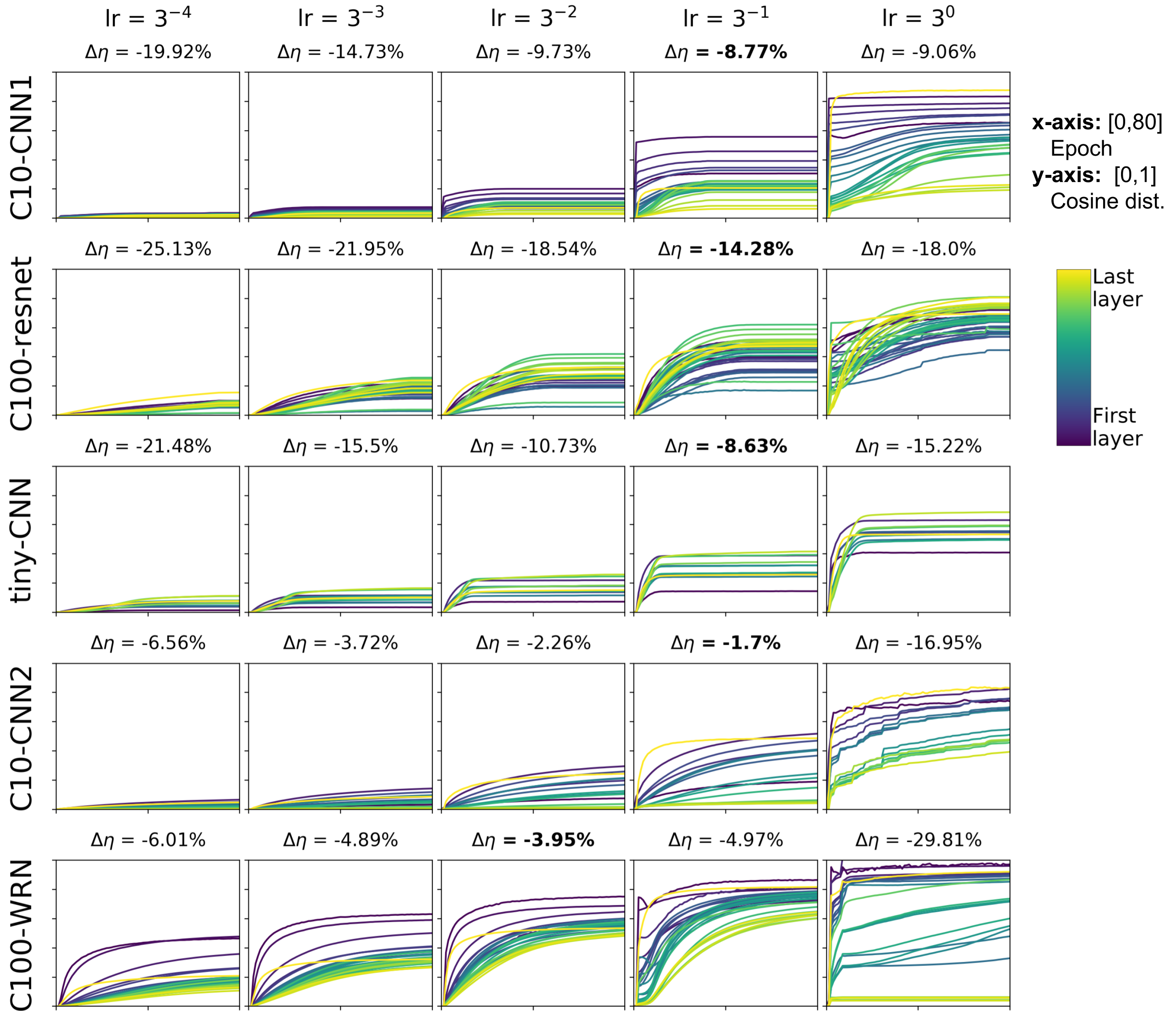}
\end{center}
\caption{Layer rotation curves and the corresponding test accuracies generated by vanilla SGD with different learning rates. Colour code, axes and $\Delta\eta$ computation are the same as in Figure \ref{fig:exploration_curves}. The influence of the learning rate parameter on generalization is consistent with our rule of thumb (larger layer rotations $\rightarrow$ better generalization), except for cases with abrupt layer rotations. We further show in Figure \ref{fig:lr_further_analysis} -Supplementary Material- that these abrupt layer rotations do not translate in improvements of the loss. Moreover, despite extensive learning rate tuning, SGD induces test performances that are significantly below Layca's optimal configuration (cfr. $5^{th}$ column of Figure \ref{fig:exploration_curves}). This is also in accordance with our rules of thumb, since SGD does not seem to be able to generate layer rotations that reach a cosine distance of 1.}
\label{fig:lr_analysis}
\end{figure}

Figure \ref{fig:lr_analysis} displays the layer rotation curves and test accuracies generated by different learning rate configurations during vanilla SGD training on the five tasks of table \ref{tab:experiments}. We observe that test accuracy increases for larger layer rotations (consistent with our rule of thumb) until a tipping point where it starts to decrease (inconsistent with our rule of thumb). Interestingly, these problematic cases also correspond to cases with extremely abrupt layer rotations that do not translate in improvements of the training loss (cfr. Figure \ref{fig:lr_further_analysis} in Supplementary Material). These observations thus highlight an expected yet important condition for our rules of thumb to hold true: the monitored layer rotations should coincide with actual training (i.e. improvements on the loss function). 

A second interesting observation is that the layer rotation curves obtained with vanilla SGD are far from the ideal scenario disclosed in Section \ref{sec:Exploration}, where the majority of the layers' weights reached a cosine distance of 1 from their initialization. In accordance with our rules of thumb, SGD also reaches considerably lower test performances than Layca. A more extensive tuning of the learning rate (over 10 logarithmically spaced values) did not help SGD to solve its two systematic problems: 1) layer rotations are not uniform and 2) the layers' weights stop rotating before reaching a cosine distance of 1.

\subsection{Analysis of SGD and weight decay} \label{sec:WDanalysis}
\begin{figure*}[!h]
\begin{center}
\includegraphics[width=.9\linewidth]{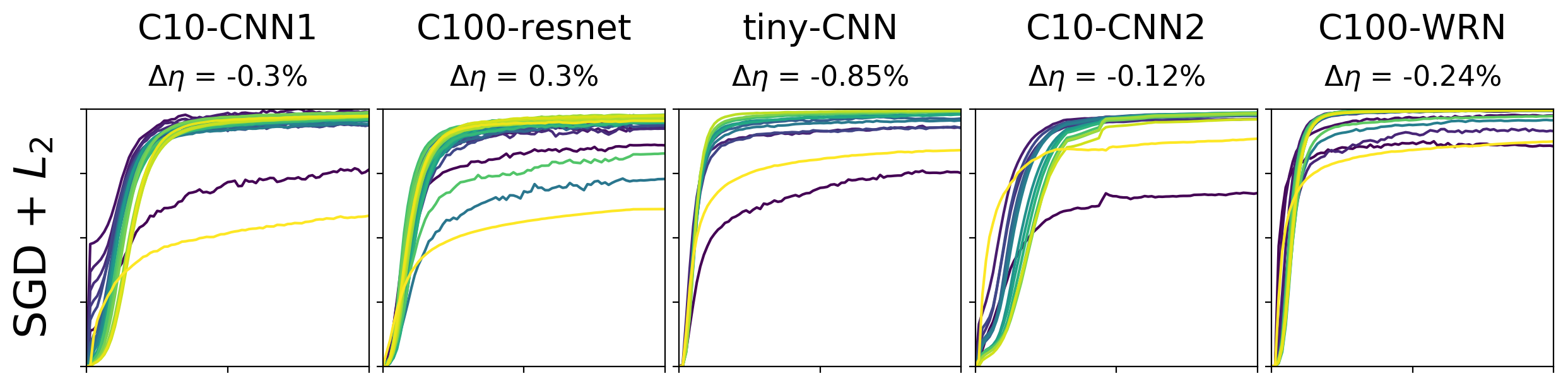}
\end{center}
\caption[Caption for LOF]{Layer rotation curves and the corresponding test accuracies generated by SGD with weight decay. Colour code, axes and $\Delta\eta$ computation are the same as in Figure \ref{fig:exploration_curves}. The application of weight decay during SGD training enables layer rotations that reach a cosine distance of 1 and leads to test performances comparable to Layca's optimal configuration (cfr. $5^{th}$ column of Figure \ref{fig:exploration_curves}).} 
\label{fig:SGD_analysis}
\end{figure*}

Several papers have recently shown that, in batch normalized networks, the regularization effect of weight decay was caused by an increase of the effective learning rate \citep{VanLaarhoven2017,Hoffer2018,Zhang2019}. More generally, reducing the norm of weights increases the amount of rotation induced by a given training step. It is thus interesting to see how weight decay affects layer rotations, and if its impact on generalization is coherent with our rule of thumb. Figure \ref{fig:SGD_analysis} displays, for the 5 tasks, the layer rotation curves generated by SGD when combined with weight decay (in this case, equivalent to $L_2$-regularization). We observe that weight decay solves SGD's problems ( cfr. Section \ref{sec:SGDanalysis}): all layers' weights reach a cosine distance of 1 from their initialization, and the resulting test performances are on par with the ones obtained with Layca.

This experiment not only provides important supplementary evidence for our rules of thumb, but also novel insights around weight decay's regularization ability in deep nets: weight decay seems to be key for enabling large layer rotations (weights reaching a cosine distance of 1 from their initialization) during SGD training. Since the same behaviour can be achieved with tools that control layer rotation rates (cfr. Layca), without an extra parameter to tune, our results could potentially lead weight decay to disappear from the standard deep learning toolkit.


\newpage

\subsection{Analysis of learning rate warmups} \label{sec:WarmUpanalysis}
We've seen in Section \ref{sec:SGDanalysis} that during SGD training, high learning rates could generate abrupt layer rotations at the very beginning of training that do not improve the training loss. In this section, we investigate if these unstable layer rotations could be the reason why learning rate warmups are sometimes necessary when using high learning rates \cite{He2016,Goyal2018}. For this experiment, we use a deeper network that notoriously requires warmups for training: ResNet-110 \cite{He2016}. The network is trained on the CIFAR-10 dataset with standard data augmentation techniques. We use a warmup strategy that starts at a 10 times smaller learning rate and linearly increases to reach the final learning rate in a specified number of epochs.

\begin{figure*}[!h]
\begin{center}
\includegraphics[width=.95\linewidth]{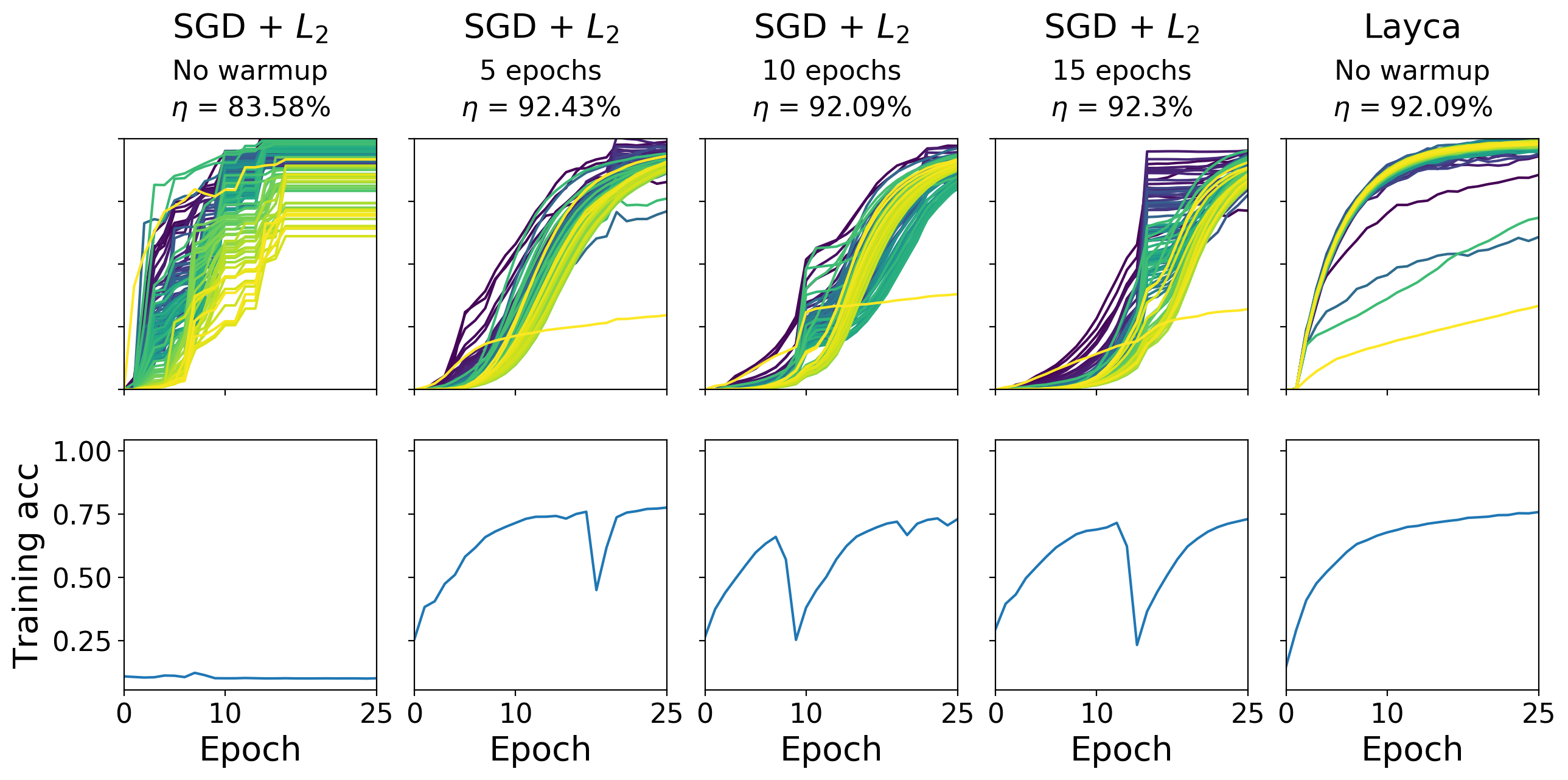}
\end{center}
\caption{Layer rotation and training curves obtained when using different warmup durations (0,5,10 or 15 epochs) during the training of ResNet-110 on CIFAR-10 with high learning rates ($3^{-1}$). The curves are shown for the 25 first epochs only -out of 200. $\eta$ is the final test accuracy (for final training accuracies, see Supplementary Material \ref{sec:trainingerrors}). We observe that SGD generates unstable layer rotations that consistently translate into inabilities to improve the training accuracy. Using warmups drastically reduce these instabilities. Layca doesn't generate instabilities thanks to its controlling ability, and reaches high generalization performance (and large layer rotations) without the need for warmups (using a learning rate of $3^3$.}
\label{fig:warmup_analysis}
\end{figure*}

Figure \ref{fig:warmup_analysis} displays the layer rotation and training curves when training with a high learning rate ($3^{-1}$) and different warmup durations (0,5,10 or 15 epochs of warmup). We observe that without warmup, SGD generates unstable layer rotations and training accuracy doesn't improve before the 25th epoch. Using warmups brings significant improvements: a $75\%$ training accuracy is reached after 25 epochs, with only some instabilities in the training curves -that again are synchronized with abrupt layer rotations. Finally, we also use Layca for training (with a $3^3$ learning rate). Thanks to its controlling ability, Layca doesn't suffer from SGD's instabilities in terms of layer rotation. We observe that this enables Layca to achieve large layer rotations and good generalization performance without the need for warmups.

\subsection{Analysis of adaptive gradient methods} \label{sec:ADGanalysis}
The recent years have seen the rise of adaptive gradient methods in the context of machine learning (\textit{e.g.} RMSProp \citep{Tieleman2012}, Adagrad \citep{Duchi2011}, Adam \citep{Kingma2015}). The key element distinguishing adaptive gradient methods from their non-adaptative equivalents is a parameter-level tuning of the learning rate based on the statistics of each parameter's partial derivative. Initially introduced for improving training speed, \citep{Wilson2017} observed that these methods also had a considerable impact on generalization. Since these methods affect the rate at which individual parameters change, they might also influence the rate at which layers change (and thus layer rotations). To emphasize this claim, we first observe to what extent the parameter-level learning rates of Adam vary across layers.

\begin{figure*}[ht]
\begin{center}
\begin{subfigure}[t]{0.7\linewidth}
\includegraphics[width=1.\linewidth]{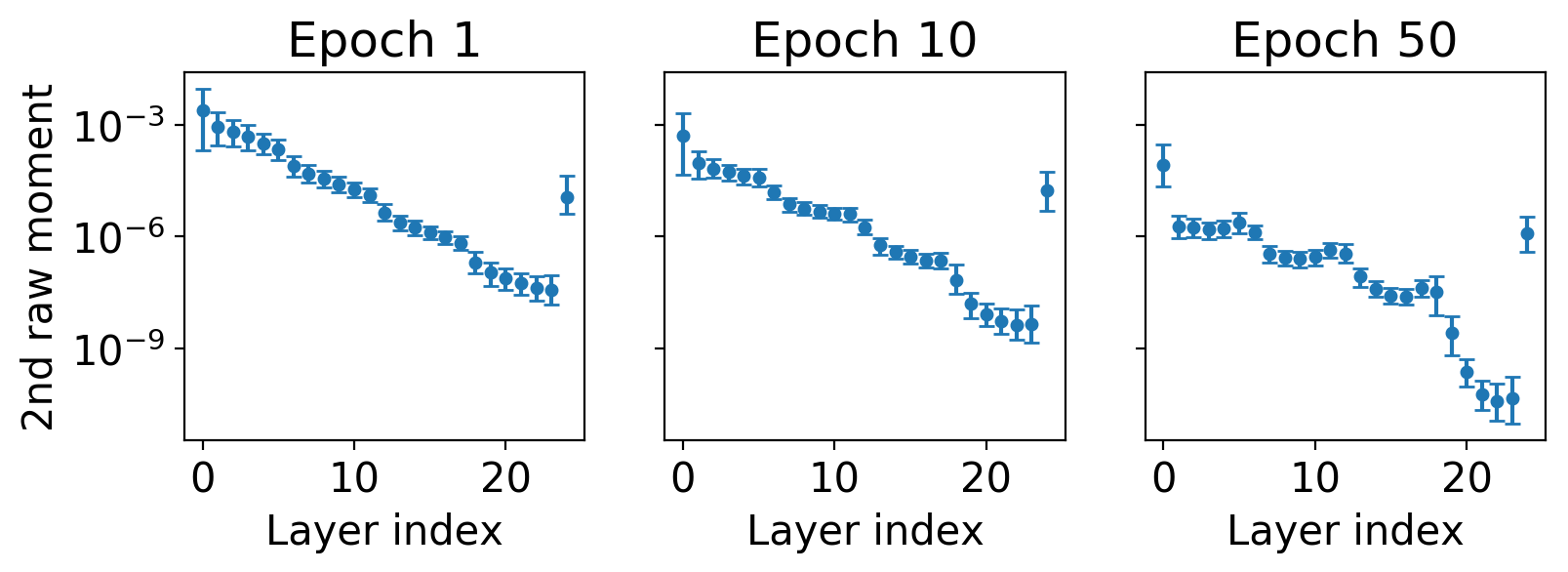}
\end{subfigure}
\end{center}
\caption{Adam's parameter-wise estimates of the second raw moment (uncentered variance) of the gradient during training on C10-CNN1, described for each layer separately through the $10^{th}$, $50^{th}$ and $90^{th}$ percentiles (represented by the lower bar, the bullet point, and the upper bar respectively for each layer index). The results provide evidence that the parameter-level statistics used by adaptive gradient methods vary mostly between layers and negligibly inside layers.}
\label{fig:adam_analysis}
\end{figure*}

We monitor Adam's estimate of the second raw moment when training on the C10-CNN1 task. The estimate is computed by:
$$v_t = \beta _2\cdot v_{t-1} + (1-\beta _2) \cdot g_t ^2 $$
where $g_t$ and $v_t$ are vectors containing respectively the gradient and the estimates of the second raw moment at training step $t$, and $\beta _2$ is a parameter specifying the decay rate of the moment estimate. While our experiment focuses on Adam, the other adaptive methods (RMSprop, Adagrad) also use statistics of the squared gradients to compute parameter-level learning rates. Figure \ref{fig:adam_analysis} displays the $10^{th}$, $50^{th}$ and $90^{th}$ percentiles of the moment estimations, for each layer separately, as measured at the end of epochs 1, 10 and 50. The conclusion is clear: the estimates vary much more across layers than inside layers. This suggests that adaptive gradient methods might have a drastic impact on layer rotations.

\textbf{Adaptive gradient methods can reach SGD's generalization ability with Layca}

\begin{figure*}[!h]
\begin{center}
\includegraphics[width=.9\linewidth]{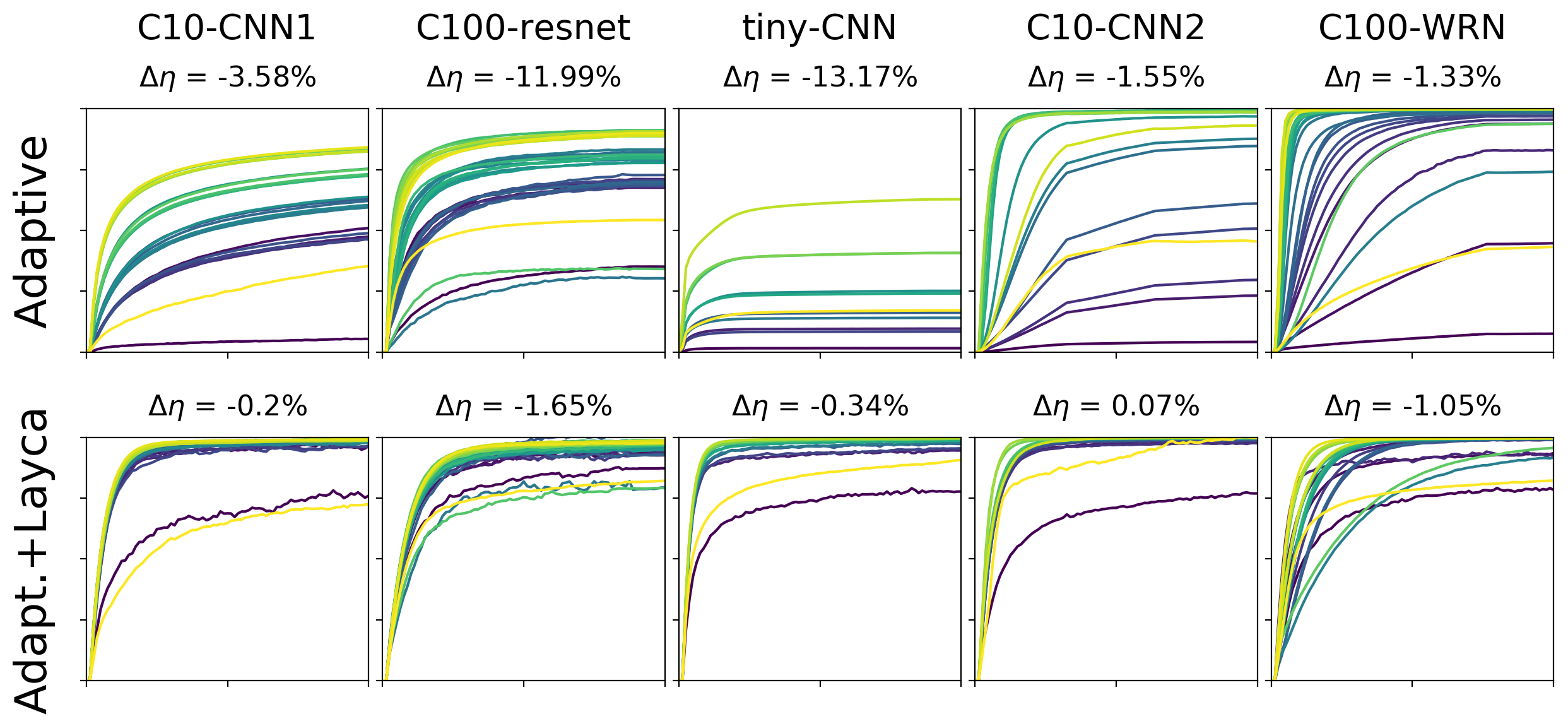} 
\end{center}
\caption{Layer rotation curves and the corresponding test accuracies generated by adaptive gradient methods (RMSProp, Adam, Adagrad, RMSProp+$L_2$ and Adam+$L_2$ respectively for each task/column) without ($1^{st}$ line) and with ($2^{nd}$ line) control of layer rotation with Layca. Colour code, axes and $\Delta\eta$ computation are the same as in Figure \ref{fig:exploration_curves}. In the first line, we observe an overall worse generalization ability compared to Layca's optimal configuration (cfr. $5^{th}$ column of Figure \ref{fig:exploration_curves}) -despite extensive learning rate tuning, together with small and/or non-uniform layer rotations (in accordance with our rule of thumb). When Layca is used on top of adaptive methods to control layer rotation (second line), adaptive methods can reach test accuracies on par with SGD + weight decay.} 
\label{fig:AGM_analysis_curves}
\end{figure*}

Since adaptive gradient methods probably affect layer rotations, we will verify if their influence on generalization is coherent with our rule of thumb. Figure \ref{fig:AGM_analysis_curves} ($1^{st}$ line) provides the layer rotation curves and test accuracies obtained when using adaptive gradient methods to train the 5 tasks described in Table \ref{tab:experiments}. We observe an overall worse generalization ability compared to Layca's optimal configuration and small and/or non-uniform layer rotations. We also observe that the layer rotations of adaptive gradient methods are considerably different from the ones induced by SGD (cfr. Figure \ref{fig:lr_analysis}). For example, adaptive gradient methods seem to induce larger rotations of the last layers' weights, while SGD usually favors rotation of the first layers' weights. Could these differences explain the impact of parameter-level adaptivity on generalization in deep learning? In Figure \ref{fig:AGM_analysis_curves} ($2^{nd}$ line), we show that when Layca is used on top of adaptive methods (to control layer rotation), adaptive methods can reach test accuracies on par with SGD + weight decay. Our observations thus suggest that adaptive gradient methods' poor generalization properties are due to their impact on layer rotations. Moreover, the results again provide supplementary evidence for our rule of thumb.

\textbf{SGD can achieve adaptive gradient methods' training speed with Layca}

\begin{figure*}[!h]
\begin{center}
\includegraphics[width=.99\linewidth]{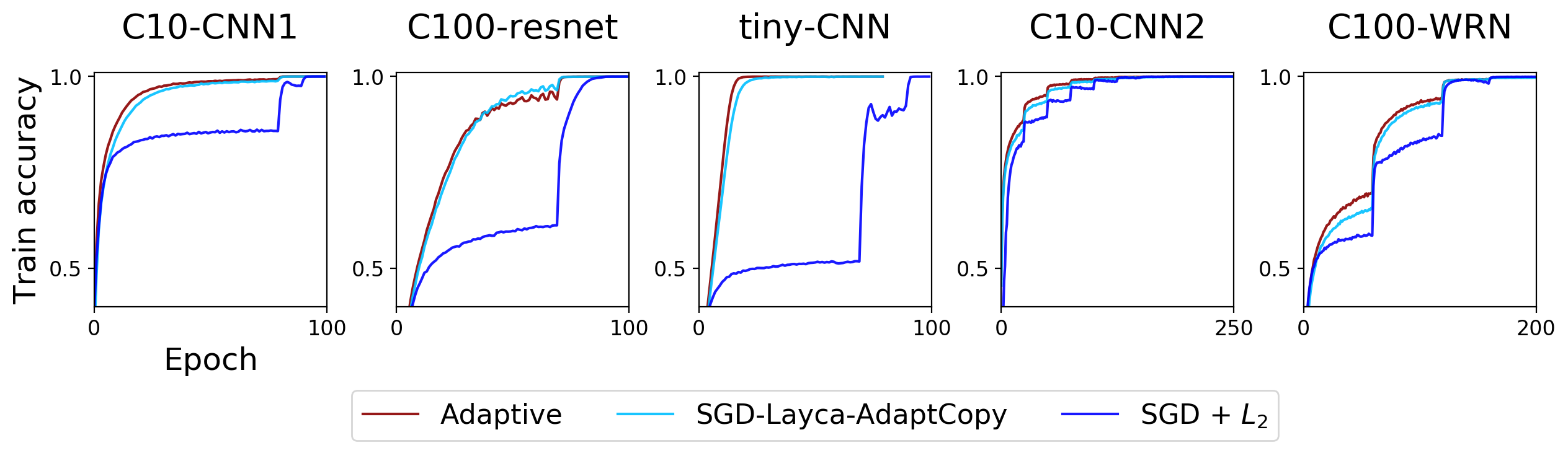} 
\end{center}
\caption[Caption for LOF]{Training curves for the 5 tasks of Table \ref{tab:experiments} with adaptive gradient methods (RMSProp, Adam, Adagrad, RMSProp+$L_2$ and Adam+$L_2$ respectively for each task/column), SGD+weight decay and SGD-Layca-AdaptCopy. During training with SGD-Layca-AdaptCopy, Layca is used to reproduce the layer rotations generated by an adaptive gradient method on the same task\protect\footnotemark. We observe that this training procedure (which doesn't perform parameter-level adaptivity) achieves the same improvements in training speed as adaptive gradient methods.}
\label{fig:AGM_analysis_histories}
\end{figure*}

We've seen that the negative impact of adaptive gradient methods on generalization was largely due to their influence on layer rotations. Could layer rotations also explain their positive impact on training speed? To test this hypothesis, we recorded the layer rotation rates emerging from training with adaptive gradient methods, and reproduced them during SGD training with the help of Layca. We then observe if this SGD-Layca optimization procedure (that doesn't perform parameter-level adaptivity) could achieve the improved training speed of adaptive gradient methods. Figure \ref{fig:AGM_analysis_histories} shows the training curves during training of the 5 tasks of Table \ref{tab:experiments} with adaptive gradient methods, SGD+weight decay and SGD-Layca-AdaptCopy (which copies the layer rotation rates of adaptive gradient methods). While adaptive gradient methods train significantly faster than SGD+weight decay, we observe that their training curves are nearly indistinguishable from SGD-Layca-AdaptCopy. Our study thus suggests that adaptive gradient methods impact on both generalization and training speed is due to their influence on layer rotations. This result relativizes the importance of parameter-level adaptivity in deep learning applications, suggesting that layer-level adaptivity is what matters most. The influence of layer rotation on training speed is further studied in Supplementary Material \ref{sec:convergence}.

\footnotetext{When copying the layer rotations of Adam with SGD+Layca, a similar momentum scheme had to be used. We call this SGD variant SGD\textunderscore AMom, and describe it further in Supplementary Material \ref{sec:SGDAMom}.}

\newpage
\section{How to interpret layer rotations?} \label{sec:future}
The previous sections of this paper demonstrate the remarkable consistency and explanatory power of layer rotation's relation with generalization in deep learning. The fundamental character of layer rotation we observe experimentally clashes however with the complete lack of theory or intuitions to support these observations. In this section, we provide a preliminary experiment and discussion to initiate a reflection on how to relate layer rotations to more established concepts in machine learning whose role during learning we can grasp more easily. 

We use a toy experiment to visualize how layer rotation affects the features learned by a network. We train a 1 hidden layer MLP (784 neurons) on a reduced MNIST dataset (1000 samples per class, to increase overparameterization). This toy network has the advantage of having intermediate features that are easily visualized: the weights associated to hidden neurons live in the same space as the input images. Starting from an identical initialization, we train the network with four different learning rates (using Layca), leading to four different layer rotation configurations that all reach $100\%$ training accuracy but different generalization abilities (in accordance with our rule of thumb). 

\begin{figure*}[!h]
\begin{center}
\includegraphics[width=1.\linewidth]{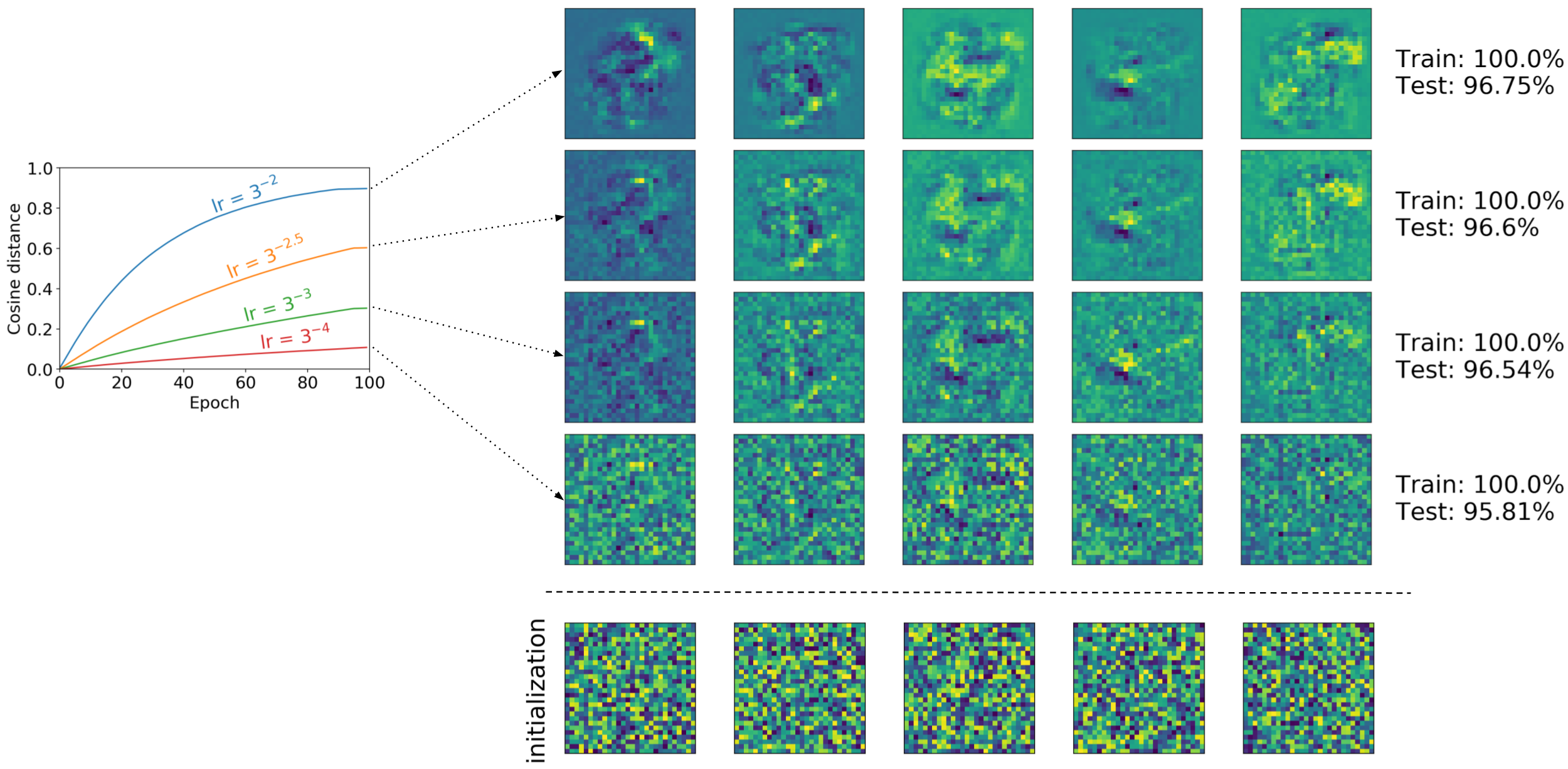} 
\end{center}
\caption{Visualization of the impact of layer rotation on intermediate features. A 1 hidden layer MLP (784 neurons) is trained on a reduced MNIST dataset (1000 samples per class). Starting from an identical initialization, the network is trained with four different learning rates (using Layca), leading to four different layer rotation configurations that all reach $100\%$ training accuracy but different generalization abilities (in accordance with our rule of thumb). The learned intermediate features (associated to 5 randomly selected neurons) are visualized for the different layer rotation configurations. The results suggest that layer rotation does not affect which features are learned, but rather to what extent they have been learned during the training process.}
\label{fig:feature_quality}
\end{figure*}

Figure \ref{fig:feature_quality} displays the features obtained by the different layer rotation configurations (for 5 randomly selected hidden neurons). This visualization unveils a remarkable phenomenon: \textbf{layer rotation does not seem to affect \textit{which} features are learned, but rather \textit{to what extent} they have been learned during the training process.} The larger the layer rotation, the more prominent the features -and the less retrievable the initialization. Ultimately, for a layer rotation close to 1, the final weights of the network got rid of all remnants of the initialization. This connection between layer rotation and the degree to which features have been learned suggest a novel interpretation of this papers' results: perfectly learning intermediate features would not be necessary to reach $100\%$ training accuracy (probably because of overparameterization), but training procedures that are able to do so anyway lead to better generalization performance.

\section{Conclusion}
This paper contains extensive empirical evidence that layer rotations constitute a remarkably powerful indicator of generalization performance. The consistency (a rule of thumb that is widely applicable), simplicity (a network-independent optimum) and explanatory power (novel insights around widely used techniques) of their relation with generalization suggest that layer rotations are tightly connected to a fundamental aspect of deep neural network training. We initiate the quest for this aspect through a preliminary experiment that reveals a connection between layer rotations and the degree by which intermediate features have been learned during the training process. We look forward to the future investigations emerging from these observations. We also hope that our (publicly available) tools for monitoring and controlling layer rotation will reduce the current struggle of practitioners to optimize hyperparameters.

\newpage

\subsubsection*{Acknowledgements}
Thanks to the anonymous NeurIPS and ICLR reviewers for their helpful feedback on previous versions of this paper. Thanks to the organizers of the ICML 2019 workshop Deep Phenomena for enabling a live presentation of this work to the community.

Special thanks to the reddit r/MachineLearning community for helping outsiders to stay up to date with the last discoveries and discussions of our fast moving field.

Simon and Christophe are Research Fellows of the Fonds de la Recherche Scientifique – FNRS, which provided funding for this work.

\bibliography{refs}
\setcitestyle{plain}
\bibliographystyle{plain}

\newpage

\appendix

\section*{Supplementary Material}
The supplementary material of this paper is divided into two sections. Section \ref{sec:SuppResults} contains supplementary results, which are not essential for the main message of the paper but could be useful for researchers interested in pursuing our line of work. Section \ref{sec:SuppInfo} contains supplementary information about the experimental procedures used in the paper.

\section{Supplementary results} \label{sec:SuppResults}

\subsection{Test accuracies for supplementary $\alpha$ and $\rho (0)$ configurations}
\begin{figure*}[!h]
\begin{center}
\includegraphics[width=.75\linewidth]{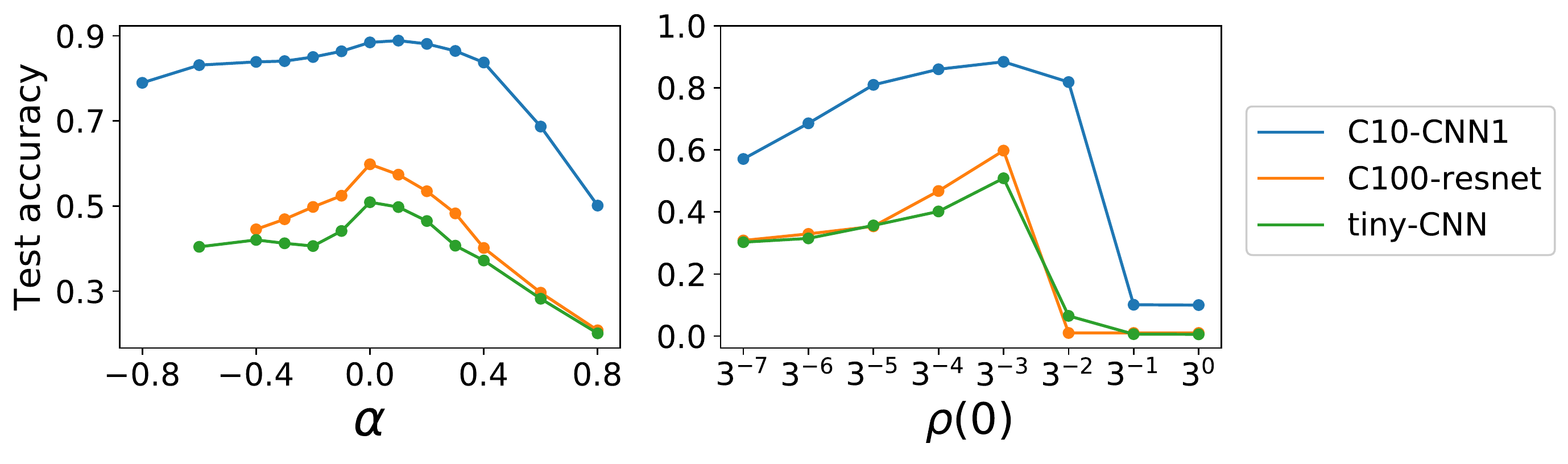} 
\end{center}
\caption{Provides test accuracies for a larger set of $\alpha$ and $\rho (0)$ values than the ones presented in \ref{fig:exploration_curves}, for the first three tasks of Table \ref{tab:experiments}.}
\label{fig:exploration_accs}
\end{figure*}

Cfr. Figure \ref{fig:exploration_accs}.

\subsection{Further analysis of high learning rates}
\begin{figure*}[!h]
\begin{center}
\includegraphics[width=.95\linewidth]{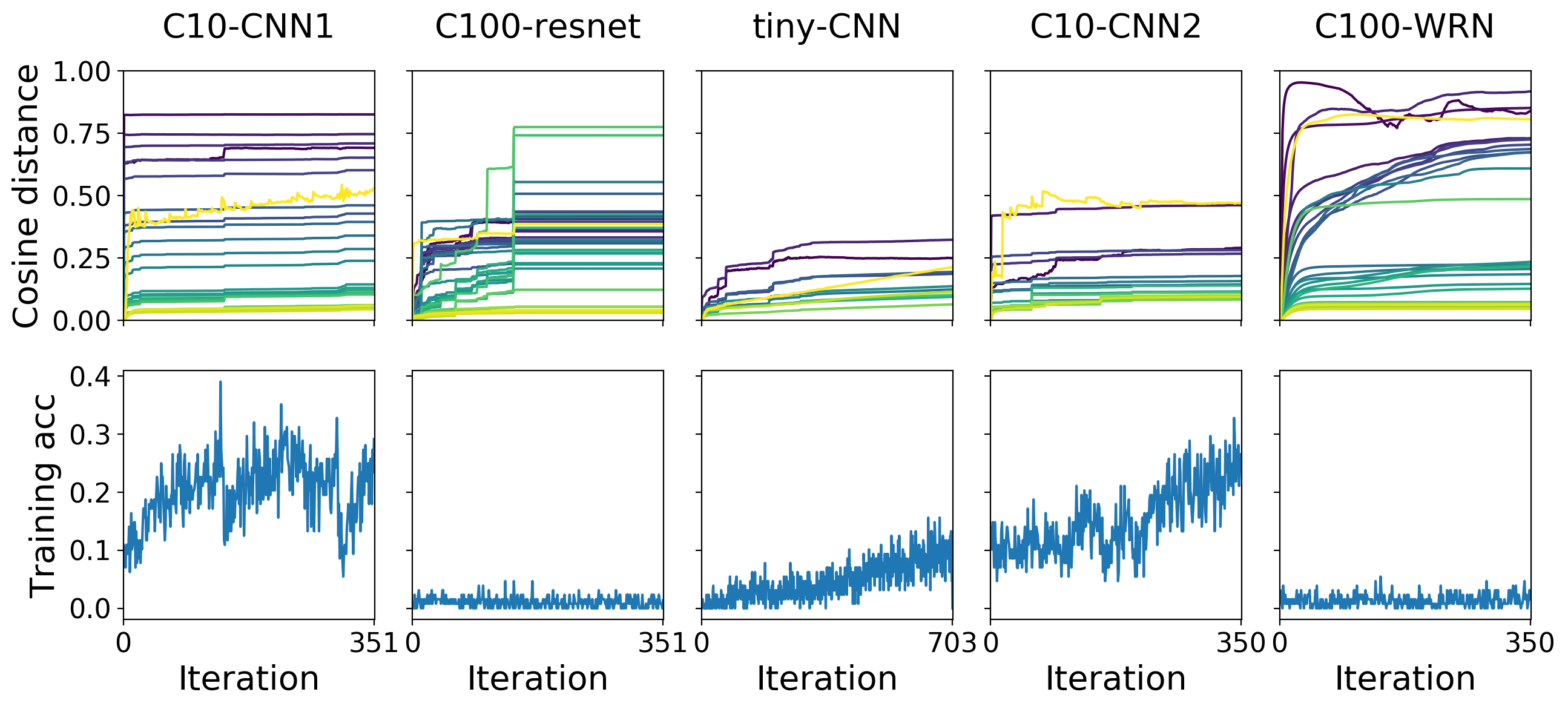}
\end{center}
\caption{Layer rotation and training curves during the first epoch of SGD training with high learning rates (cfr. Figure \ref{fig:lr_analysis}). The visualization reveals large layer rotations sometimes performed in a single iteration. Importantly, these iterations do not induce improvements in training accuracy. Such configurations escape the scope of our rule of thumb.}
\label{fig:lr_further_analysis}
\end{figure*}

Figure \ref{fig:lr_analysis} reveals unstable layer rotations when using high learning rates with SGD. Figure \ref{fig:lr_further_analysis} takes a closer look at this phenomenon by plotting layer rotation and training curves at iteration-level (instead of epoch-level) precision during the first epoch of training. The visualization reveals large layer rotations sometimes performed in a single iteration. Importantly, these iterations do not induce improvements in training accuracy. Such configurations escape the scope of our rule of thumb.

\subsection{How layer rotation influences training speed} \label{sec:convergence}
\begin{figure}[!h]
\begin{center}
\includegraphics[width=.95\linewidth]{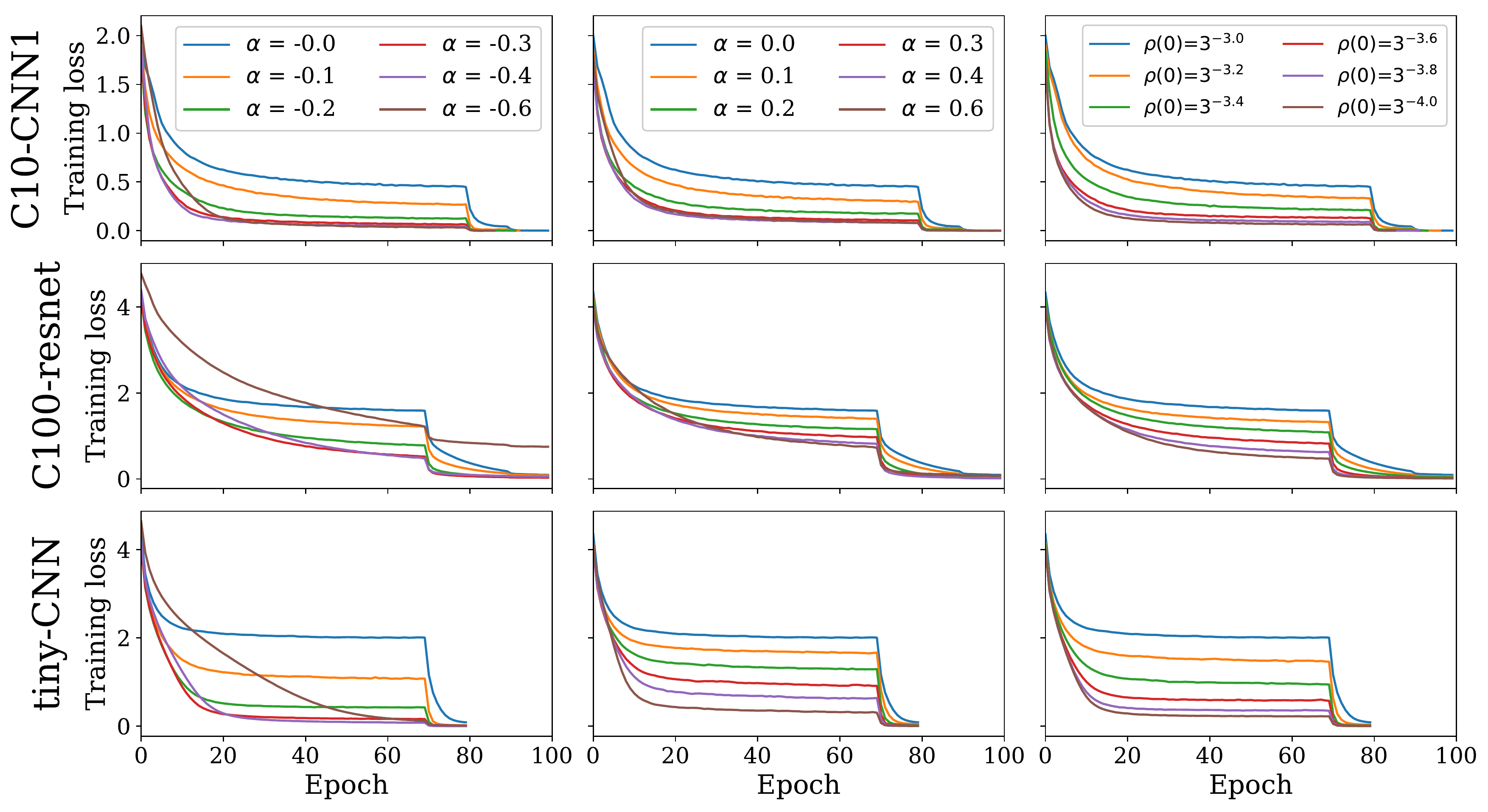} 
\end{center}
\caption{Loss curves obtained for different $\alpha$ and $\rho (0)$ values on the first three tasks of Table \ref{tab:experiments}, using Layca for training ($\alpha$ and $\rho (0)$ configurations are specified in the legends -there is one per per column). The visualizations unveil a remarkable phenomenon: the more uniform or the larger the layer rotation rates, the higher the plateaus in which the loss gets stuck into. The sudden drop at epoch 70 corresponds to a reduction of the global learning rate by a factor $5$ or $10$.}
\label{fig:convergence}
\end{figure}

While generalization is the main focus of our paper, we observed through our experiments that layer rotation rates also influenced the loss curves of our models in a remarkable way. Figure \ref{fig:convergence} depicts the loss curves obtained for different values of $\alpha$ and $\rho (0)$ on the first three tasks of Table \ref{tab:experiments}. It appears that the larger or the more uniform the layer rotation rates are, the higher the plateaus in which loss curves get stuck into. This curious phenomenon again emphasizes the fundamental character of layer rotation. We have no explanation of this behaviour however.

Following our rule of thumb, this result also suggests that high plateaus are additional indicators of good generalization performance. This is consistent with the systematic occurrence of high plateaus in the loss curves of state of the art networks \citep{He2016,Zagoruyko2016} (which usually use SGD with weight decay). 

\subsection{All operations of Layca are not always necessary in practice.} \label{sec:SuppLARS}
The 4 main operations of Layca are repeated in Algorithm \ref{alg:Layca_bis}. The first operation projects the step on the space orthogonal to the current weights of the layer. Having a step orthogonal to the current weights is necessary for operation 2 to normalize the rotation performed during the update. However, since a layer typically has more than thousands of parameters (\textit{i.e.} has a lot of dimensions), the step proposed by an optimizer has a high probability of being approximately orthogonal to the current weights. Explicitly orthogonalizing the step and the weights through operation 1 is thus potentially redundant. Operation 4 keeps the norm of weights fixed during the whole training process. This operation prevents the weights from increasing too much (the first three operations lead the norm of weights to increase at every training step), which causes numerical problems. However, this operation is not fundamental for controlling the layer rotation rates.

\begin{figure}[!h]
\begin{center}
\includegraphics[width=.6\linewidth]{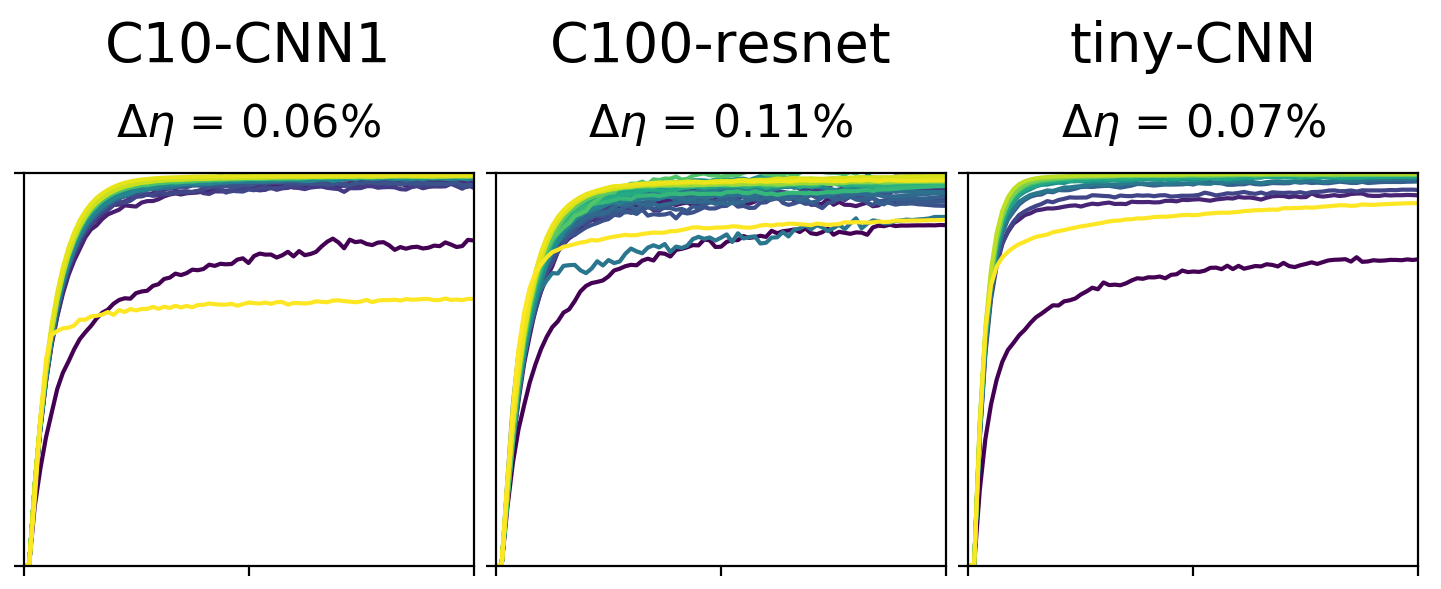} 
\end{center}
\caption{Layer rotation curves and the corresponding test accuracies generated by LARS with $\alpha = 0$ and $\rho (0) = 3^{-3}$. Colour code, axes and $\Delta\eta$ computation are the same as in Figure \ref{fig:exploration_curves}. Although not performing operations 1 and 4 of Algorithm \ref{alg:Layca_bis}, LARS seems to control layer rotation rates as well as Layca. Indeed, the layer rotation curves are indistinguishable from the ones in the $5^{th}$ column of Figure \ref{fig:exploration_curves}, and the test accuracies are nearly identical.}
\label{fig:LARS}
\end{figure}

We experimented with a sub-version of Layca that does not perform Layca's operations 1 and 4. Interestingly, the resulting algorithm is equivalent to $\text{NG}_{\text{adap}}$ and LARS introduced by \citep{Yu2017} and \citep{Ginsburg2018} respectively. Both works reported improved test performance when using this algorithm. Figure \ref{fig:LARS} shows the layer rotation curves and associated test accuracies when applying LARS (or equivalently, $\text{NG}_{\text{adap}}$) on the C10-CNN1, C100-resnet and tiny-CNN tasks.\footnote{While the norm of each layer's weight vector was not fixed by LARS, we still had to limit the amount of norm increase per training step to prevent numerical errors. We limited it to $0.0001$ times the initial norm of each layer's weight vector.} The layer rotation rate configuration parameters are $\alpha = 0$ and $\rho (0) = 3^{-3}$. We observe that this configuration also induces large layer rotations, and that the test accuracies are on par with Layca. This observation indicates that operations 1 and 4 of Layca can be removed in at least some practical applications.

\begin{algorithm}[!h]
   \caption{Main operations of Layca (cfr. Algorithm \ref{alg:layca}). We've noticed that in practice, operations 1 and 4 are not strictly necessary for controlling layer rotation rates.}
   \label{alg:Layca_bis}
\begin{algorithmic}
   \State $s^0_t,..., s^{L-1}_t  = \text{getStep}(o,w^0_t,..., w^{L-1}_t)$ \noindent\hspace{6pt} (get the updates of the selected optimizer)
   \For{l=0 {\bfseries to} L-1}
   \State $s^l_t \leftarrow s^l_t - \frac{(s^l_t \cdot w^l_t) w^l_t}{w^l_t \cdot w^l_t}$ \noindent\hspace{24pt} (1: project step on space orthogonal to $w^l_t$)
   \State $s^l_t \leftarrow \frac{s^l_t \parallel w^l_t\parallel _2}{\parallel s^l_t\parallel _2}$                 \noindent\hspace{50pt} (2: rotation-based normalization)
   \State $w^l_{t+1} \leftarrow w^l_t + \rho_l(t) s^l_t$ \noindent\hspace{18pt} (3: perform update)
   \State $w^l_{t+1} \leftarrow w^l_{t+1} \frac{\parallel w^l_{0}\parallel _2}{\parallel w^l_{t+1}\parallel _2}$ \noindent\hspace{16pt} (4: project weights back on sphere)
   \EndFor
\end{algorithmic}
\end{algorithm}

\section{Supplementary information}\label{sec:SuppInfo}

\subsection{Visualizing the $\alpha$ parameter.} \label{sec:alphaviz}
\begin{figure*}[!h]
\begin{center}
\includegraphics[width=0.9\linewidth]{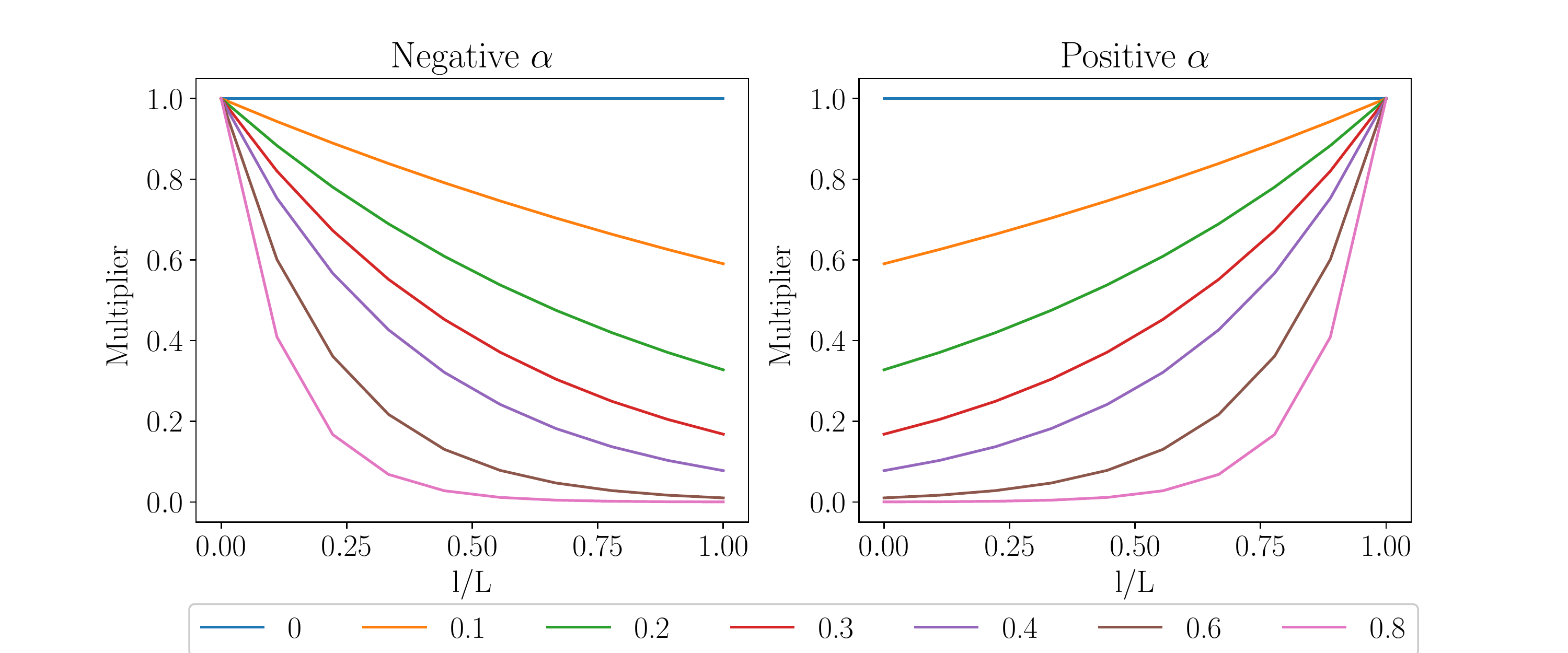}
\end{center}
\caption{Visualization of the prioritization schemes as parametrized by $\alpha$ (cfr. Section \ref{sec:Exploration}). The colours of the lines represent the absolute value of $\alpha$. Illustration is separated for faster rotation of the first layers (negative $\alpha$ values) and of the last layers (positive $\alpha$ values). The layer-wise learning rate multipliers (y-axis) depend on the layer's location in the network (x-axis), which is represented by the layer index $l$ (in forward pass ordering) divided by the number of layers $L$.}
\label{fig:alpha}
\end{figure*}

The $\alpha$ parameter is used in Section \ref{sec:Exploration} to characterize the layer prioritization schemes used during training. While the specific parametrization is provided in Equation \ref{eq:alpha}, Figure \ref{fig:alpha} provides a graphical illustration of it.

\subsection{Learning rate decay schemes} \label{sec:lrDecay}
Our work uses standard learning rate decay schemes, as follows:\\
\begin{itemize}
\item C10-CNN1: 100 epochs and a reduction of the learning rate by a factor 5 at epochs 80, 90 and 97
\item C100-resnet: 100 epochs and a reduction of the learning rate by a factor 10 at epochs 70, 90 and 97
\item tiny-CNN: 80 epochs and a reduction of the learning rate by a factor 5 at epoch 70
\item C10-CNN2: 250 epochs and a reduction of the learning rate by a factor 5 at epochs 100, 170, 220
\item C100-WRN: 250 epochs and a reduction of the learning rate by a factor 5 at epochs 100, 170, 220
\end{itemize}
The only exceptions are C10-CNN2 and C100-WRN training with SGD+weight decay and with adaptive methods, where the learning rate decay schemes are the ones used in their original implementation or in \citep{Wilson2017}.

\subsection{Training errors associated to the layer rotation curves.} \label{sec:trainingerrors}
In Figures \ref{fig:exploration_curves}, \ref{fig:lr_analysis}, \ref{fig:SGD_analysis}, \ref{fig:warmup_analysis} and \ref{fig:AGM_analysis_curves}, the test accuracies corresponding to each layer rotation curves visualization are provided. While it is briefly mentioned that training accuracy is close to perfect in most cases, Tables \ref{tab:train_exploration}, \ref{tab:train_lr}, \ref{tab:train_SGD}, \ref{tab:train_warmup} and \ref{tab:train_AGM} provide the exact values for completeness.

\begin{table*}[!h]
  \caption{Train accuracies associated to Figure \ref{fig:exploration_curves}}
  \label{tab:train_exploration}
  \centering
  \begin{tabular}{l|ccccc|}
     & $\alpha = 0.6$     & $\alpha = -0.6$  & $\rho (0) = 3^{-5}$ &  $\rho (0) = 3^{-4}$ & Best\\ \hline 
    C10-CNN1     & \rule{0pt}{2.6ex}$100\%$  & $99.99\%$   & $100\%$ & $100\%$ &  $99.99\%$ \\[3pt]
    C100-resnet  & $82.09\%$  & $99.54\%$ &$99.87\%$ & $99.99\%$  &$99.75\%$  \\[3pt]
    tiny-CNN     & $99.98\%$ & $99.95\%$ &$99.97\%$ &$99.97\%$ &$98.91\%$ \\[3pt]
    C10-CNN2  & $100\%$  & $99.94\%$ &$99.99\%$ & $99.99\%$  &$99.97\%$  \\[3pt]
    C100-WRN  & $99.88\%$  & $99.91\%$ &$99.97\%$ & $99.99\%$  &$99.96\%$  \\[2pt]
    \hline
  \end{tabular}
\end{table*}

\begin{table*}[!h]
  \caption{Train accuracies associated to Figure \ref{fig:lr_analysis}}
  \label{tab:train_lr}
  \centering
  \begin{tabular}{l|ccccc|}
     & $lr = 3^{-4}$     & $lr = 3^{-3}$  & $lr = 3^{-2}$ &  $lr = 3^{-1}$ & $lr = 3^{0}$\\ \hline 
    C10-CNN1     & \rule{0pt}{2.6ex}$100\%$  & $100\%$   & $100\%$ & $100\%$ &  $100\%$ \\[3pt]
    C100-resnet  & $87.8\%$  & $100\%$ &$100\%$ & $100\%$  &$99.7\%$  \\[3pt]
    tiny-CNN     & $100\%$ & $100\%$ &$100\%$ &$100\%$ &$100\%$ \\[3pt]
    C10-CNN2  & $99.8\%$  & $99.9\%$ &$100\%$ & $100\%$  &$83.7\%$  \\[3pt]
    C100-WRN  & $100\%$  & $100\%$ &$100\%$ & $100\%$  &$57.4\%$  \\[2pt]
    \hline
  \end{tabular}
\end{table*}

\begin{table*}[!h]
  \caption{Train accuracies associated to Figure \ref{fig:SGD_analysis}}
  \label{tab:train_SGD}
  \centering
  \begin{tabular}{l|ccccc|}
     & C10-CNN1 & C100-resnet  & tiny-CNN &  C10-CNN2 & C100-WRN \\ \hline
    SGD + $L_2$    & \rule{0pt}{2.6ex}$100\%$  & $100\%$   & $100\%$ & $100\%$ &  $100\%$ \\[2pt]
    \hline
  \end{tabular}
\end{table*}

\begin{table*}[!h]
  \caption{Train accuracies associated to Figure \ref{fig:warmup_analysis}}
  \label{tab:train_warmup}
  \centering
  \begin{tabular}{l|ccccc|}
     & No warmup & 5 epochs  & 10 epochs &  15 epochs & Layca-No warmup \\ \hline
        & \rule{0pt}{2.6ex}$96.67\%$  & $99.76\%$   & $99.85\%$ & $99.68\%$ &  $99.85\%$ \\[2pt]
    \hline
  \end{tabular}
\end{table*}

\begin{table*}[!h]
  \caption{Train accuracies associated to Figure \ref{fig:AGM_analysis_curves}}
  \label{tab:train_AGM}
  \centering
  \begin{tabular}{l|ccccc|}
     & C10-CNN1 & C100-resnet  & tiny-CNN &  C10-CNN2 & C100-WRN \\ \hline 
    Adaptive methods    & \rule{0pt}{2.6ex}$100\%$  & $100\%$   & $100\%$ & $100\%$ &  $99.9\%$ \\[3pt]
    Adaptive + Layca  & $100\%$  & $99.7\%$ &$99.2\%$ & $100\%$  &$100\%$ \\[2pt]
    \hline
  \end{tabular}
\end{table*}

\subsection{Momentum scheme used by SGD\textunderscore AMom and Adam.} \label{sec:SGDAMom}
SGD\textunderscore AMom was designed for Section \ref{sec:ADGanalysis}, as a non-adaptive equivalent of Adam. In particular, SGD\textunderscore AMom uses the same momentum scheme as Adam:
\begin{eqnarray*}
v_t &=& m \cdot v_{t-1} + (1-m)\cdot g_t \\
w_t &=& w_{t-1} - \rho \cdot v_t
\end{eqnarray*}
where $g_t$ is the gradient at step $t$, $\rho$ the learning rate, $m$ the momentum parameter.


%
%

\end{document}